\pdfoutput=1
\documentclass[11pt]{article}
\usepackage{EMNLP2022}
\usepackage{times}
\usepackage{latexsym}
\usepackage[T1]{fontenc}
\usepackage[utf8]{inputenc}
\usepackage{microtype}
\usepackage{inconsolata}
\usepackage[]{graphicx,indentfirst}
\usepackage{subfigure}
\usepackage{multirow}
\usepackage{amsmath}
\usepackage{amssymb}
\usepackage{tabulary}

\title{Bi-Directional Iterative Prompt-Tuning for Event Argument Extraction}

\author{Lu Dai$^1$,\quad Bang Wang$^1$,\quad Wei Xiang $^1$,\quad Yijun Mo$^2$ \\ %
$^1$School of Electronic Information and Communications, \\ Huazhong University of Science and Technology, Wuhan, China \\
$^2$School of Computer Science and Technology, \\ Huazhong University of Science and Technology, Wuhan, China \\
\texttt{\{dailu18, wangbang, xiangwei, moyj\}@hust.edu.cn}  \\
}

\begin{document}
\maketitle
\begin{abstract}
Recently, prompt-tuning has attracted growing interests in event argument extraction (EAE). However, the existing prompt-tuning methods have not achieved satisfactory performance due to the lack of consideration of entity information. In this paper, we propose a bi-directional iterative prompt-tuning method for EAE, where the EAE task is treated as a cloze-style task to take full advantage of entity information and pre-trained language models (PLMs). Furthermore, our method explores event argument interactions by introducing the argument roles of contextual entities into prompt construction. Since template and verbalizer are two crucial components in a cloze-style prompt, we propose to utilize the role label semantic knowledge to construct a semantic verbalizer and design three kinds of templates for the EAE task. Experiments on the ACE 2005 English dataset with standard and low-resource settings show that the proposed method significantly outperforms the peer state-of-the-art methods. Our code is available at https://github.com/HustMinsLab/BIP.
\end{abstract}

\section{Introduction}\label{Sec:Introduction}
As a key step of event extraction, event argument extraction refers to identifying event arguments with predefined roles. For example, for an "Attack" event triggered by the word "fired" in the sentence "\emph{\underline{Iraqis} have \textbf{fired} sand \underline{missiles} and \underline{AAA} at \underline{aircraft}}", EAE aims to identify that "Iraqis", "missiles", "AAA" and "aircraft" are event arguments with the "Attacker", "Instrument", "Instrument" and "Target" roles, respectively.

\begin{figure}[h]
    \centering
    \subfigure[Fine-Tuning for EAE]{
    \includegraphics[width=0.47\textwidth]{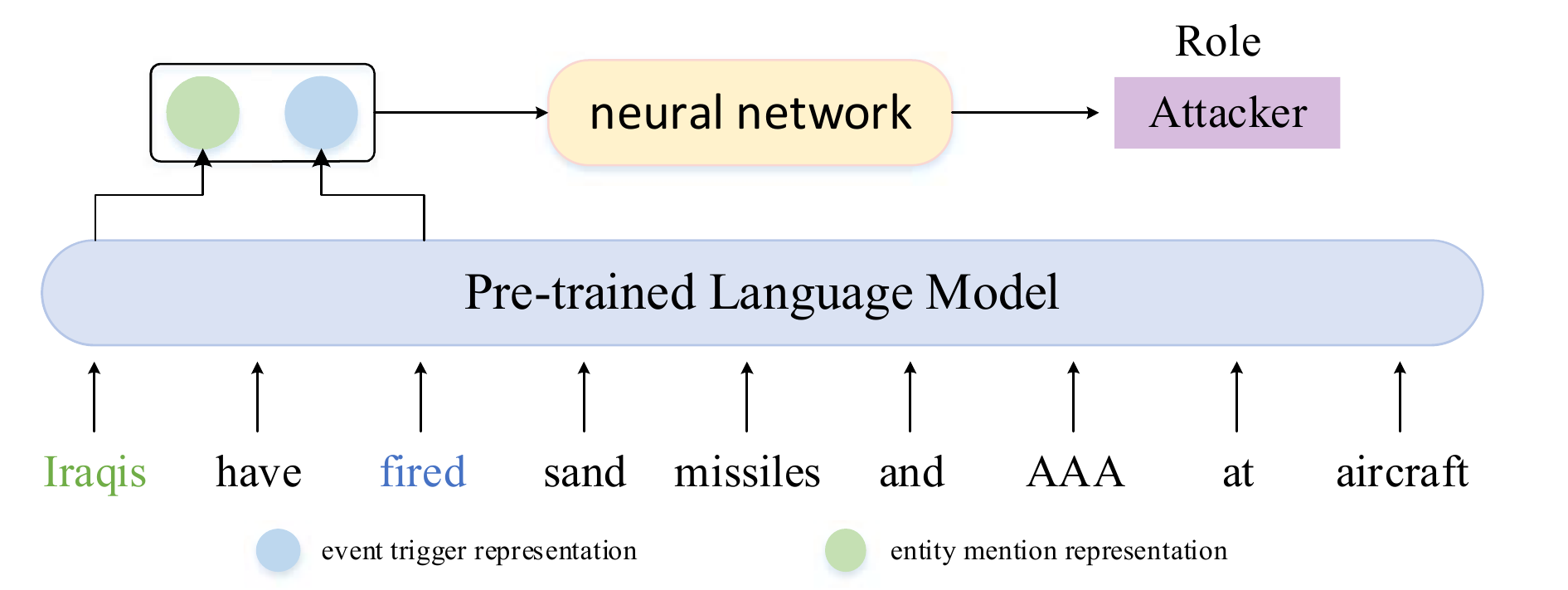}}
    \subfigure[Prompt-Tuning for EAE.]{
    \includegraphics[width=0.47\textwidth]{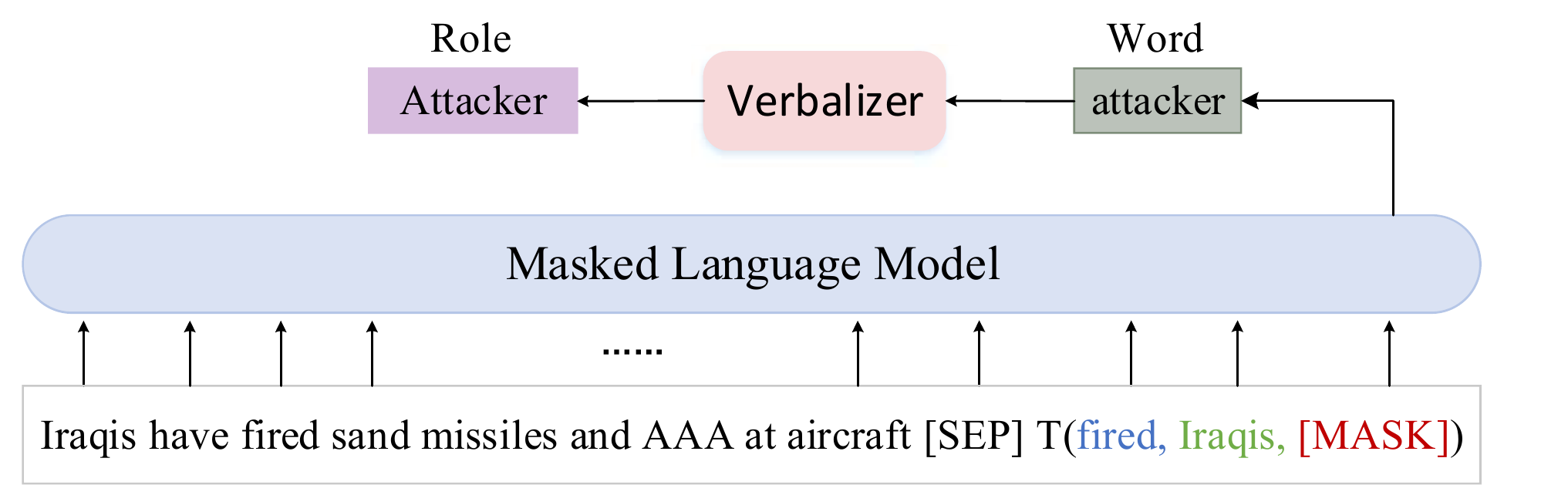}}

    \caption{Illustration of fine-tuning and prompt-tuning methods for predicting the argument role of the entity mention "Iraqis" in the event triggered by the word "fired".}\label{Fig:PF}
\end{figure}

\par
In order to exploit the rich linguistic knowledge contained in pre-trained language models, fine-tuning methods have been proposed for EAE. The paradigm of these methods is to use a pre-trained language model to obtain semantic representations, and then feed these representations into a well-designed neural network to extract event arguments. For example in Figure~\ref{Fig:PF}(a), an event trigger representation and an entity mention representation are first obtained through a pre-trained language model, and then input to a designed neural network, such as hierarchical modular network~\citep{wang-etal-2019-hmeae} and syntax-attending transformer network~\citep{ma-etal-2020-resource}, to determine the argument role that the entity mention plays in the event triggered by the trigger. However, there is a significant gap between the EAE task and the objective form of pre-training, resulting in the poor utilization of the prior knowledge in PLMs. Additionally, fine-tuning methods heavily depend on extensive annotated data and perform poorly in low-resource data scenarios.

\par
To bridge the gap between the EAE task and the pre-training task, prompt-tuning methods~\citep{li-etal-2021-document,ma-etal-2022-prompt,naacl2022degree,liu-etal-2022-dynamic} recently have been proposed to formalize the EAE task into a more consistent form with the training objective of generative pre-trained language models. These methods achieve significantly better performance than fine-tuning methods in low-resource data scenarios, but not as good as the state-of-the-art fine-tuning method ONEIE~\citep{lin-etal-2020-joint} in high-resource data scenarios.

\par
To achieve excellent performance in both low-resource and high-resource data scenarios, we leverage entity information to model EAE as a cloze-style task and use a masked language model to handle the task. Figure~\ref{Fig:PF}(b) shows a typical cloze-style prompt-tuning method for EAE. The typical prompt-tuning method suffers from two challenges: (i) The typical human-written verbalizer~\citep{schick-schutze-2021-just} is not a good choice for EAE. The human-written verbalizer is to manually assign a label word to each argument role. For example in Figure~\ref{Fig:PF}(b), we choose the "attacker" as the label word of "Attacker" role. However, an argument role may have different definitions in different types of events. For example, the "Entity" role refers to "the voting agent" and "the agents who are meeting" in the "Elect" and "MEET" events, respectively. (ii) Event argument interactions are not explored. Existing work~\citep{Sha:et.al:2018,xiangyu-etal-2021-capturing,ma-etal-2022-prompt} has demonstrated the usefulness of event argument interactions for EAE. For the "Attack" event triggered by the word "fired" in Figure~\ref{Fig:PF}, given that "missiles" is an "Instrument", it is more likely to correctly classify "AAA" into the "Instrument" role.

\par
In this paper, we propose a \emph{bi-directional iterative prompt-tuning} (BIP) method to alleviate the aforementioned challenges. To capture argument interactions, a forward iterative prompt and a backward iterative prompt are constructed to utilize the argument roles of contextual entities to predict the current entity's role. For the verbalizer, we redefine the argument role types and assign a virtual label word to each argument role, where the initial representation of each virtual label word is generated based on the semantic of the argument role. In addition, we design three kind of templates: hard template, soft template, and hard-soft template, which are further discussed in the experimental section. Extensive experiments on the ACE 2005 English dataset show that the proposed method can achieve the state-of-the-art performance in both low-resource and high-resource data scenarios.

\section{Related Work}\label{Sec:RelatedWork}
In this section, we review the deep learning methods for event argument extraction and prompt-tuning methods for natural language processing.

\subsection{Event Argument Extraction}\label{Subsec:EAE}
Early deep learning methods use various neural networks to capture the dependencies in between event triggers and event arguments to extract event arguments, such as convolutional neural network (CNN)-based models~\citep{chen-etal-2015-event}, recurrent neural network (RNN)-based models~\citep{nguyen-etal-2016-joint-event,Sha:et.al:2018} and graph neural networks (GNN)-based models~\citep{liu-etal-2018-jointly,dai2021event}. As pre-trained language models have been proven to be powerful in language understanding and generation~\citep{devlin-etal-2019-bert,liu2019roberta,lewis-etal-2020-bart}, some PLM-based methods have been proposed to extract event arguments. These methods can be divided into two categories: fine-tuning and prompt-tuning ones.

\par
Fine-tuning methods aim to design a variety of neural network models to transfer pre-trained language models to EAE task. According to the modeling manner of EAE task, existing fine-tuning work can be further divided into three groups: classification-based methods~\citep{wang-etal-2019-hmeae,wadden-etal-2019-entity,lin-etal-2020-joint,ma-etal-2020-resource,xiangyu-etal-2021-capturing}; machine reading comprehension-based methods~\citep{du-cardie-2020-event,li-etal-2020-event,liu-etal-2020-event}; generation-based methods~\citep{paolini2020structured,lu-etal-2021-text2event}. Prompt-tuning methods aim to design a template to provide useful prompt information for pre-trained language models to extract event arguments~\citep{li-etal-2021-document,ma-etal-2022-prompt,naacl2022degree,liu-etal-2022-dynamic}. For example, \citet{li-etal-2021-document} create a template for each event type based on the event ontology definition and model the EAE task as the conditional text generation. This method acquires event arguments by comparing the designed template with the generated natural language text. \citet{naacl2022degree} improve the method of \citet{li-etal-2021-document} by replacing the non-semantic placeholder tokens in the designed template with words with role label semantics.

\subsection{Prompt-tuning}\label{Subsec:PT}
The core of prompt-tuning is to transform the given downstream task into a form that is consistent with a training task of the pre-trained language models~\citep{liu2021pre}. As prompt-tuning makes better use of prior knowledge contained in pre-trained language models, this new paradigm is beginning to become popular in NLP tasks and has achieved promising performance~\citep{seoh-etal-2021-open,han2021ptr,cui2021template,hou-etal-2022-inverse,hu-etal-2022-knowledgeable,chen2022knowprompt}. For example, \citet{cui2021template} use candidate entity spans and entity type label words to obtain templates, and recognize entities based on the pre-trained generative language model's score for each template. \citet{hu-etal-2022-knowledgeable} convert the text classification task to a masked language modeling problem by predicting the word filled in the "[\texttt{MASK}]" token, and propose a knowledgeable verbalizer to map the predicted word into a label. \citet{chen2022knowprompt} consider the relation extraction problem as a cloze task and use the relation label semantic knowledge to initialize the virtual label word embedding for each relation label.

\section{Model}\label{Sec:Model}
In this section, we first introduce the problem description of event argument extraction and the overall framework of our bi-directional iterative prompt-tuning method, then explain the details of designed semantical verbalizer, three different templates, and model training.

\begin{figure*}[t]
    \centering
    \includegraphics[width=1.0\textwidth]{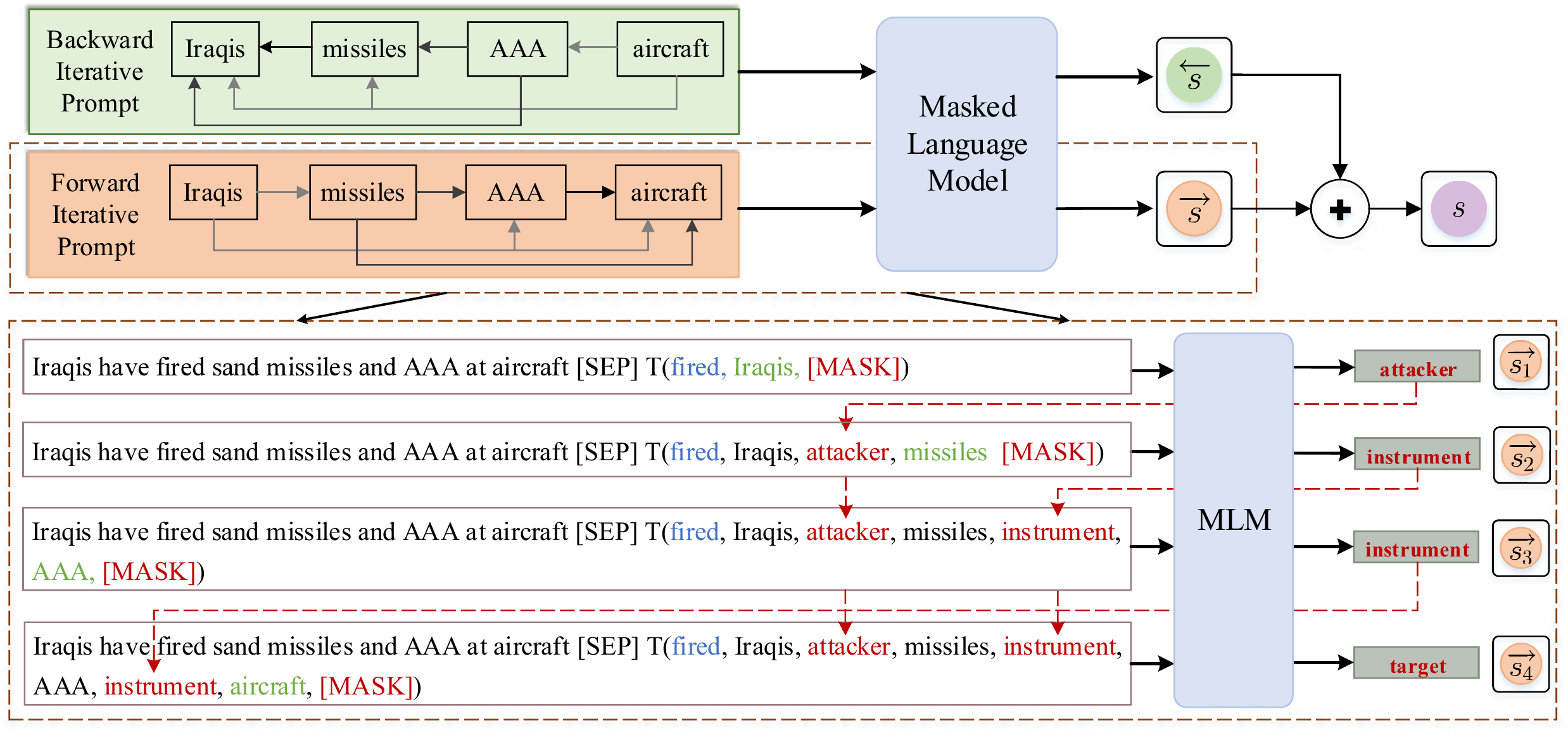}
    \caption{The overall architecture of our bi-directional iterative prompt-tuning method shown with an example predicting argument roles of "Iraqis", "missiles", "AAA", and "aircraft" in the "Attack" event triggered by "fired", where blue font represents the given trigger and green font represents the given entity.}\label{Fig:model}
\end{figure*}

\subsection{Problem Description}\label{Subsec:Overview}
As the most common ACE dataset provides entity mention, entity type and entity coreference information, we use these entity information to formalize event argument extraction into the argument role prediction problem of entities. The detailed problem description is as follow: \textit{Given a sentence $S$, an event trigger $t$ with event type, and $n$ entities $\{e_1,e_2,...,e_n\}$, the goal is to predict the argument role of each entity in the event triggered by $t$ and output a set of argument roles $\{r_1,r_2,...,r_n\}$.}

In this paper, the argument role prediction problem is casted as a cloze-style task through a template $T(\cdot)$ and verbalizer. For the trigger $t$ and entity $e_i$, a template $T(t, e_i, [\texttt{MASK}])$ is constructed to query the argument role that the entity $e_i$ plays in the event triggered by $t$. For example in Figure~\ref{Fig:PF}(b), the template $T(fired, Iraqis, [\texttt{MASK}])$ can be set as "\emph{For the \underline{attack} event triggered by the fired, the \underline{person}, Iraqis, is [\texttt{MASK}]}", where "\underline{attack}" represents the event type of the trigger "fired" and "\underline{person}" represents the entity type of the entity "Iraqis". Then the input of the $i$-th entity $e_i$ is:
\begin{equation}
x_i = S \ [\texttt{SEP}] \ T(t, e_i, [\texttt{MASK}]).
\end{equation}

\par
The verbalizer is a mapping from the label word space to the argument role space. Let $l_j$ denote the label word that is mapped into the role $r_j$, the confidence score that the $i$-th entity is classified as the $j$-th role type is:
\begin{equation}
s_{ij} = C_i([\texttt{MASK}]=l_j),
\end{equation}
where $C_i$ is the output of a pre-trained masked language model at the masked position in $x_i$, i.e. the confidence score of each word in the dictionary filled in the [\texttt{MASK}] token.

\subsection{Overall Framework}\label{Subsec:Overview}
Figure~\ref{Fig:model} presents the overall architecture of our bi-directional iterative prompt-tuning method, consisting of a forward iterative prompt (FIP) and a backward iterative prompt (BIP). The forward iterative prompt predicts the argument role of each entity iteratively from left to right until argument roles of all entities are obtained. For example in Figure~\ref{Fig:model}, the order of entities is "$Iraqis \rightarrow missiles \rightarrow AAA \rightarrow aircraft$".

\par
In order to utilize the predicted argument role information to classify the current entity into the correct role, we introduce the argument roles of the first $i$-$1$ entities into the template of the $i$-th entity. The template of the $i$-th entity in the forward iterative prompt can be represented as:
\begin{equation}
FIP(e_i)= T(t, e_1, \overrightarrow{l_1}, ..., e_{i-1}, \overrightarrow{l_{i-1}}, e_i, [\texttt{MASK}]),
\end{equation}
where $\overrightarrow{l_j}$ is the role label word of the $j$-th entity predicted by the forward iterative prompt. For example in Figure~\ref{Fig:model}, $\overrightarrow{l_1}$ is the word "attacker". Then the confidence score distribution of the $i$-th entity over all argument roles in the forward iterative prompt can be computed by
\begin{equation}
\overrightarrow{\mathbf{s}_i} = MLM(S \ [\texttt{SEP}] \ FIP(e_i)).
\end{equation}
$\overrightarrow{l_i}$ is the word corresponding to the argument role with the highest value in $\overrightarrow{\mathbf{s}_i}$.

\par
Similarly, the backward iterative prompt predicts the argument role of each entity in a right-to-left manner. The argument role confidence score distribution of the $i$-th entity in the backward iterative prompt can be computed by:
\begin{gather}
BIP(e_i)= T(t, e_n, \overleftarrow{l_n}, ..., e_{i+1}, \overleftarrow{l_{i+1}}, e_i, [\texttt{MASK}]), \\
\overleftarrow{\mathbf{s}_i} = MLM(S \ [\texttt{SEP}] \ BIP(e_i)).
\end{gather}

Then we can obtain the final argument role confidence score distribution of the $i$-th entity by
\begin{equation}
\mathbf{s}_i = \overrightarrow{\mathbf{s}_i} + \overleftarrow{\mathbf{s}_i}.
\end{equation}
Finally, the argument role label with the highest score is chosen as the role prediction result.

\subsection{Semantical Verbalizer}\label{Subsec:Label}
To tackle the problem that an argument role may have different definitions in different types of events, we reconstruct the set of argument role types and design a semantical verbalizer. Specifically, we further divide the argument role that participates in multiple types of events into multiple argument roles that are specific to event types. For example, the "Entity" role is divided into "Elect:Entity", "Meet:Entity", and etc. Since the "Place" role has the same meaning in all types of events, we do not consider to divide it.

\par
For each new argument role, the semantical verbalizer constructs a virtual word to represent the role and initializes the representation of the virtual word with the semantic of the argument role. Let a $m$-word sequence $\{q_{i1},q_{i2},...,q_{i,m}\}$ denote the semantic description of the argument role $r_i$, the initial representation of the label word $l_i$ that is mapped into the role $r_i$ can be computed by:
\begin{equation}
\mathbf{E}(l_i) = \frac{1}{m} \sum\limits^{m}_{j=1} \mathbf{E}(q_{ij}),
\end{equation}
where $\mathbf{E}$ is the word embedding table of a pre-trained masked language model.

\par
For redefined argument roles, different argument roles may have the same semantics, such as "Appeal:Adjudicator" and "Sentence:Adjudicator". Therefore, it is easy to misclassify the entity with "Appeal:Adjudicator" role into the "Sentence:Adjudicator" role. In order to solve the problem, we use the event structure information to extract arguments. For an event with the "Appeal" type, its role label can only be "Appeal:Defendant", "Appeal:Adjudicator" and "Appeal:Plaintiff".

\subsection{Templates}\label{Subsec:Templates}
\begin{figure*}[t]
    \centering
    \includegraphics[width=1.0\textwidth]{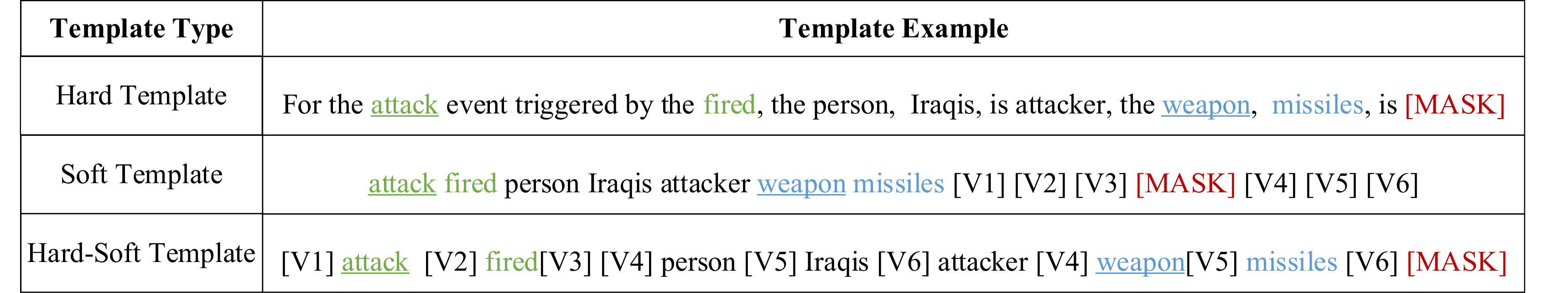}
    \caption{Examples of three different templates with trigger "fired" and entity "missiles", where green font represents the given trigger, green underlined font represents the text span of event type, blue font represents the given entity, blue underlined font represents the text span of entity type.}\label{Fig:Temp}
\end{figure*}
To take full advantage of event type, trigger, and entity information, the designed template should contain event types, triggers, entity types, and entity mentions. Since some entity types and event types are not human-understandable words, such as "PER" and "Phone-Write", we need to convert each entity (event) type into a human-understandable text span. For example, we use "person" and "'written or telephone communication" as the text spans for "PER" and "Phone-Write" respectively.

\par
Let $M_i = \{\varepsilon_{i1},\varepsilon_{i2},...,\varepsilon_{id}\}$ denote the entity mention set of the $i$-th entity, the word sequence of the $i$-th entity can be represented as:
\begin{equation}
\hat{e}_{i} = \varepsilon_{i1}\ or\ \varepsilon_{i2}\ or\ ...\ or\ \varepsilon_{id}.
\end{equation}
We use $w^t$ to denote the text span of event type of the given trigger and $w_i^e$ to denote the text span of the entity type of the $i$-th entity. For the given trigger $t$ and $i$-th entity $e_i$, three different templates of forward iterative prompt are designed as follows:
\begin{itemize}
  \item Hard Template: All known information are connected manually with natural language. "\textit{For the $w^t$ event triggered by the $t$, the $w_1^e$, $\hat{e}_1$, is $\overrightarrow{l_1}$, ... , the $w_{i-1}^e$, $\hat{e}_{i-1}$, is $\overrightarrow{l_{i-1}}$, the $w_i^e$, $\hat{e}_i$, is [\texttt{MASK}]}"
  \item Soft Template: Add a sequence of learnable pseudo tokens after all known information. "\textit{$w^t$ $t$ $w_1^e$ $\hat{e}_1$ $\overrightarrow{l_1}$ ... $w_{i-1}^e$ $\hat{e}_{i-1}$ $\overrightarrow{l_{i-1}}$ $w_{i}^e$ $\hat{e}_{i}$ [V1] [V2] [V3] [\texttt{MASK}] [V4] [V5] [V6]}"
  \item Hard-Soft Template: All known information are connected with learnable pseudo tokens. "\textit{[V1] $w^t$ [V2] $t$ [V3] [V4] $w_1^e$ [V5]  $\hat{e}_1$ [V6] $\overrightarrow{l_1}$, ... , [V4] $w_{i-1}^e$ [V5] $\hat{e}_{i-1}$ [V6] $\overrightarrow{l_{i-1}}$ [V4] $w_i^e$ [V5] $\hat{e}_i$ [V6] [\texttt{MASK}]}"
\end{itemize}
Pseudo tokens are represented by "[Vi]". The embedding of each pseudo token is randomly initialized and optimized during training.

\subsection{Training}\label{Subsec:Training}
During training, gold argument roles are used to generate the template of each entity in forward iterative prompt and backward iterative prompt. The optimization objective is to ensure that the masked language model can predict argument role accurately in both forward iterative prompt and backward iterative prompt. We use $\overrightarrow{p_{t,i}}$ and $\overleftarrow{p_{t,i}}$ to represent the probability of the entity $e_i$ playing each role type in the event triggered by $t$ in forward and backward iterative prompt respectively. The loss function is defined as follows:
\begin{gather}
\overrightarrow{p_{t,i}} = softmax(\overrightarrow{\mathbf{s}_i}),\ \  \overleftarrow{p_{t,i}} = softmax(\overleftarrow{\mathbf{s}_i}),  \notag \\
\mathbb{L}=-\sum\limits_{t\in \mathbb{T}}\sum\limits^{n_t}_{i=1}(\log(\overrightarrow{p_{t,i}}(\tilde{r}_{t,i}))+\log(\overleftarrow{p_{t,i}}(\tilde{r}_{t,i}))),
\end{gather}
where $\mathbb{T}$ is the event trigger set in the training set, $n_t$ is the number of entities contained in the same sentence as the event trigger $t$, and $\tilde{r}_{t,i}$ is the correct argument role of the $i$-th entity playing in the event triggered by $t$.

\section{Experiments}\label{Sec::Experiments}

\subsection{Experimental Setup}\label{Subsec:Setup}
We evaluate our proposed method on the most widely used event extraction dataset, ACE 2005 English dataset\footnote{\url{https://catalog.ldc.upenn.edu/LDC2006T06}}~\citep{Doddington:et.al:2004}. Following the previous work~\citep{wadden-etal-2019-entity,lin-etal-2020-joint,ma-etal-2022-prompt}, the dataset is pre-processed and divided into training/development/test set, where $33$ event subtypes, $7$ entity types and $22$ argument roles are considered in the processed dataset. As event argument extraction task is only focused on, we use gold entities and event triggers to conduct experiments.

\par
We use Bert-base(containing around 110 millions parameters)~\citep{devlin-etal-2019-bert} and Roberta-base(containing around 125 millions parameters)~\citep{liu2019roberta} models to predict the masked words and train each model with AdamW, where the batch size is set to $4$ and the learning rate is set to $1e$-$5$. For the low-resource setting, we generate some subsets containing $(1\%, 5\%, 10\%, 20\%, 50\%, 75\%)$ of the fulling training set in the same way as ~\citep{naacl2022degree}. In each experiment, the masked language model is trained by a subset and evaluated by the fulling development and test sets. All experiments are run on a NVIDIA Quadro P4000 GPU.

\begin{table*}[t]
\centering
\small
\begin{tabular}{|c|c|c|c|c|c|c|c|c|}
\hline
\multirow{2}{*}{PLM}  & \multirow{2}{*}{Model}  & \multirow{2}{*}{Eval}    & \multicolumn{3}{c|}{Argument Identification} & \multicolumn{3}{c|}{Role Classification} \\ \cline{4-9}
&                       &     & P & R  & F & P  & R & F \\ \hline   \hline
\multirow{6}{*}{Bert}     & HMEAE & SM   & 65.22 & 68.08  & 66.62 & 60.06  & 62.68 & 61.34 \\
                      & (EMNLP, 2019)  & FM & 73.67 & 72.70  & 73.18 & 66.86  & 65.99 & 66.42 \\ \cline{2-9}
& ONEIE         & SM   & 73.65 & 71.72  & 72.67 & 69.31  & 67.49 & 68.39 \\
& (ACL, 2020)   & FM & \textbf{79.48} & 75.77  & 77.58 & 74.89  & 71.39 & 73.09 \\ \cline{2-9}
& BERD          & SM   & 68.83 & 66.62  & 67.70 & 63.25  & 61.22 & 62.22 \\
& (ACL, 2021)   & FM & 76.01 & 71.04  & 73.55 & 69.63  & 65.26 & 67.37\\ \hline  \hline
\multirow{6}{*}{Roberta}    & HMEAE & SM   & 70.37 & 69.24  & 69.80 & 64.00  & 62.97 & 63.48 \\
                        & (EMNLP, 2019)  & FM & 76.58 & 72.55  & 74.51 & 69.49  & 65.84 & 67.62 \\ \cline{2-9}
& ONEIE         & SM   & 72.86 & 73.18  & 73.02 & 69.81  & 70.12 & 69.96 \\
& (ACL, 2020)   & FM & 78.55 & 79.12  & \underline{78.84} & \underline{75.22}  & \underline{75.77} & \underline{75.50} \\ \cline{2-9}
& BERD          & SM   & 69.03 & 69.53  & 69.28 & 63.24  & 63.70 & 63.47 \\
& (ACL, 2021)   & FM & 75.72 & 73.28  & 74.48 & 69.08  & 66.86 & 67.95\\ \hline  \hline
\multirow{4}{*}{Bart}   &DEGREE(EAE)& SM   & 70.39 & 68.95 & 69.66 & 65.77 & 64.43 & 65.10 \\
&                         (NAACL, 2022)  & FM & \underline{79.20} & 75.60 & 77.37 & 74.16 & 70.80 & 72.44 \\ \cline{2-9}
& PAIE          & SM   & 72.16 & 71.12  & 71.64 & 68.65  & 66.71 & 67.67 \\
& (ACL, 2022)   & FM & 76.75 & \underline{79.55}  & 78.13 & 72.82  & 74.22 & 73.51 \\ \hline   \hline
Bert    & BIP(our)                       &  & 75.54 & 81.29 & 78.31 &  71.60  & 77.05 & 74.23 \\ \hline
Roberta & BIP(our)                       &  & 78.17 \tiny{(-1.31)} & \textbf{86.40 \tiny{(+6.85)}} & \textbf{82.08 \tiny{(+3.24)}} & \textbf{75.26 \tiny{(+0.04)}} & \textbf{83.19 \tiny{(+7.42)}} & \textbf{79.03 \tiny{(+3.53)}} \\ \hline
\end{tabular}
\caption{\label{Tbl:Result}
Experiment results of our proposed method with hard template and baselines, where the boldface is the best results, the underline is the second best results, and results of baselines are our re-implementations. Due to the limited memory of our GPU, only base-version models are adopted to perform experiments.}
\end{table*}

\subsection{Baselines}\label{Subsec:Baselines}
Two categories of state-of-the-art methods are compared with our proposed method.

\subsubsection*{Fine-tuning Methods:}
\begin{itemize}
  \item \textbf{HMEAE}~\citep{wang-etal-2019-hmeae} is a hierarchical modular model that uses the superordinate concepts of argument roles to extract event arguments.
  \item \textbf{ONEIE}~\citep{lin-etal-2020-joint} is a neural framework that leverages global features to jointly extract entities, relations, and events. When applying ONEIE to the EAE task, we also use gold entity mentions and event triggers to extract event arguments, without considering the relations.
  \item \textbf{BERD}~\citep{xiangyu-etal-2021-capturing} is a bi-directional entity-level recurrent decoder that utilizes the argument roles of contextual entities to predict argument roles entity by entity.
\end{itemize}

\subsubsection*{Prompt-tuning Methods:}
\begin{itemize}
  \item \textbf{DEGREE(EAE)}~\citep{naacl2022degree} summarizes an event into a sentence based on a designed prompt containing the event type, trigger, and event-type-specific template. Then event arguments can be extracted by comparing the generated sentence with the event-type-specific template.
  \item \textbf{PAIE}~\citep{ma-etal-2022-prompt} is an encoder-decoder architecture, where the given context and designed event-type-specific prompt are input into the encoder and decoder separately to extract event argument spans.
\end{itemize}

\subsection{Evaluation}\label{Subsec:Evaluation}
Since we use an entity as a unit for argument role prediction, an event argument is correctly identified if the entity corresponding to the argument is predicted to be the non-None role type. The argument is further be correctly classified if the predicted role type is the same as the gold label.

\par
For the above baselines, they consider that an event argument is correctly classified only if its offsets and role type match the golden argument, which can be called "\textbf{strict match (SM)}". In order to compare our model with baselines more fairly, we use a "\textbf{flexible match (FM)}" method to evaluate these baselines, that is, an argument is correctly classified if its offsets match any of the entity mentions co-referenced with the golden argument and role type match the golden argument.

\par
Same as the previous work, the standard micro-averaged Precision(P), Recall(R), and F1-score(F1) are used to evaluate all methods.

\subsection{Overall Results}\label{Subsec:Results}
Table~\ref{Tbl:Result} compares the overall results between our model and baselines, from which we have several observations and discussions.

\par
(1) BIP(Roberta) gains the significant improvement in event argument extraction. The F1-scores of  BIP(Roberta) are more than $9\%$ higher than those of all baselines obtained by the strict match evaluation method. Even using the flexible match method to evaluate baselines, the BIP(Roberta) method also outperforms the state-of-the-art ONEIE(Roberta) by $3.24\%$ increase of F1-score in term of argument identification and $3.53\%$ increase of F1-score in term of role classification.

\par
(2) Comparing with the strict match, the flexible match achieves $5\%$ to $7\%$ F1-score improvements in term of argument identification and role classification. These results indicate that the trained argument extraction models can indeed identify the entity mention co-referenced with the golden argument as the event argument. In addition, in the actual application scenarios, we only pay attention to which entity is the event argument, not which mention in an entity is the event argument. Therefore, it is more reasonable and efficient to predict argument roles in unit of entity than entity mention.

\par
(3) Roberta-version methods outperform Bert-version methods. In particular, for our proposed BIP method, Roberta further gains $3.77\%$ and $4.8\%$ F1-score improvements on argument identification task and role classification task respectively. These improvements can be explained by Roberta using a much larger training dataset than Bert and removing the next sentence prediction task. In the following experiments, we only consider Roberta-version methods.

%

\begin{table}[t]
\centering
\begin{tabular}{lccc}
\hline
\multirow{2}{*}{Model} & \multicolumn{3}{c}{Role Classification}\\ \cline{2-4}
                        & P  & R & F1 \\ \hline
BIP(our)           & 75.26 & 83.19 & 79.03 \\ \hline
BIP(forward)       & 76.06 & 78.95 & 77.47 \\ \hline
BIP(backward)      & 75.94 & 76.61 & 76.27 \\ \hline
-BI                & 78.79 & 76.02 & 77.38 \\ \hline
-SV                & 74.79 & 78.07 & 76.39 \\ \hline
-BI-SV             & 78.19 & 74.42 & 76.25 \\
\hline
\end{tabular}
\caption{An ablation study of our proposed method.}\label{Tbl:Ablation}
\end{table}

\subsection{Ablation Study}\label{Subsec:Ablation}
Table~\ref{Tbl:Ablation} presents an ablation study of our proposed BIP method. BIP(forward) only considers the forward iterative prompt to extract event arguments. BIP(backward) only considers the backward iterative prompt. BIP-BI does not use a bi-directional iterative strategy to consider argument interactions, i.e. predicts the argument role of each entity separately. BIP-SV replaces our designed semantical verbalizer with a human-written verbalizer, where each label word is manually selected from a pre-trained language model vocabulary. BIP-BI-SV uses neither the bi-directional iterative strategy nor the semantical verbalizer. Some observations on the ablation study are as follows:

\par
(1) Compared with the method BIP, the performance of BIP(forward) and BIP(backward) is decreased by $1.56\%$ and $2.76\%$ F1-score in term of role classification, respectively. These results clearly demonstrate that the bi-directional iterative prompt-tuning can further improve the performance by comparing with one direction.

\par
(2) Comparing with the methods BIP-BI and BIP-BI-SV, the methods BIP and BIP-SV can further improve the performance of role classification in terms of $1.65\%$ and $0.14\%$ increase of F1-score, respectively. These results suggest that the bi-directional iterative strategy is useful for event argument extraction. In addition, we notice that the improvement brought by our bi-directional iterative strategy for the method BIP-BI is higher than BIP-BI-SV. This suggests that the more accurate the independent predicted argument role of each entity, the greater improvement the bi-directional iterative strategy will bring to the performance of argument extraction.

\par
(2) The methods BIP and BIP-BI are respectively outperform the methods BIP-SV and BIP-BI-SV by $2.64\%$ and $1.13\%$ F1-score in term of role classification. These results illustrate that our semantical verbalizer is more effective than a human-written verbalizer for event argument extraction.

\begin{table*}[t]
\centering
\small
\resizebox{\textwidth}{!}{
\begin{tabular}{p{3.0cm}<{\centering}|p{2cm}<{\centering}|p{2cm}<{\centering}|p{2cm}<{\centering}|p{2cm}<{\centering}|p{2cm}<{\centering}}
\hline
\multicolumn{6}{l}{Sentence 1: Swapping smiles, handshakes and hugs at a joint press appearance after talks linked to Saint Petersburg's} \\
\multicolumn{6}{l}{300th anniversary celebrations, \emph{Bush} and \emph{Putin} set out to recreate the buddy atmosphere of their previous encounters.} \\ \hline
\multicolumn{6}{l}{Event Trigger: \textbf{talks}, Event Type: \textbf{Meet}}     \\ \hline
\multicolumn{6}{l}{Extraction Results:}     \\ \hline
Entity & BIP & BIP(forward) & BIP(backward) & BIP-BI & BIP-SV\\ \hline
Bush   & Entity($\checkmark$)  & Entity($\checkmark$) & None($\times$) & Entity($\checkmark$) & Entity($\checkmark$) \\ \hline
Putin  & Entity($\checkmark$)  & Entity($\checkmark$) & None($\times$) & None($\times$) & Entity($\checkmark$) \\ \hline
\hline

\multicolumn{6}{l}{Sentence 2: Earlier Saturday, \emph{Baghdad} was again targeted, one day after a massive U.S. aerial bombardment in which} \\
\multicolumn{6}{l}{ more than 300 \emph{Tomahawk} cruise \emph{missiles} rained down on the \emph{capital}.} \\ \hline
\multicolumn{6}{l}{Event Trigger: \textbf{targeted}, Event Type: \textbf{Attack}}     \\ \hline \hline
\multicolumn{6}{l}{Extraction Results:}     \\ \hline
Entity & BIP & BIP(forward) & BIP(backward) & BIP-BI & BIP-SV\\ \hline
[Baghdad, capital]   & Place($\checkmark$)  & Place($\checkmark$) & Place($\checkmark$) & Place($\checkmark$) & Place($\checkmark$) \\ \hline
[Tomahawk, missiles]  & None($\checkmark$)  & Instrument($\times$) & None($\checkmark$) & None($\checkmark$) & None($\checkmark$) \\ \hline
\hline

\multicolumn{6}{l}{Sentence 3: Last month, the \emph{SEC} slapped fines totaling 1.4 billion dollars on 10 Wall Street \emph{brokerages} to settle charges} \\
\multicolumn{6}{l}{ of conflicts of interest between analysts and investors.} \\ \hline
\multicolumn{6}{l}{Event Trigger: \textbf{fines}, Event Type: \textbf{Fine}}     \\ \hline
\multicolumn{6}{l}{Extraction Results:}     \\ \hline
Entity & BIP & BIP(forward) & BIP(backward) & BIP-BI & BIP-SV\\ \hline
SEC   & Adjudicator($\checkmark$)  & Adjudicator($\checkmark$) & Adjudicator($\checkmark$) & Adjudicator($\checkmark$) & Entity($\times$) \\ \hline
brokerages  & Entity($\checkmark$)  & Entity($\checkmark$) & Entity($\checkmark$) & Entity($\checkmark$) & Entity($\checkmark$) \\ \hline
\hline

\end{tabular}}
\caption{Event argument extraction results by different methods. }\label{Tbl:Case2}
\end{table*}

\subsection{Low-Resource Event Argument Extraction}\label{Subsec:LowResource}
\begin{figure}[h]
    \centering
    \includegraphics[width=0.45\textwidth]{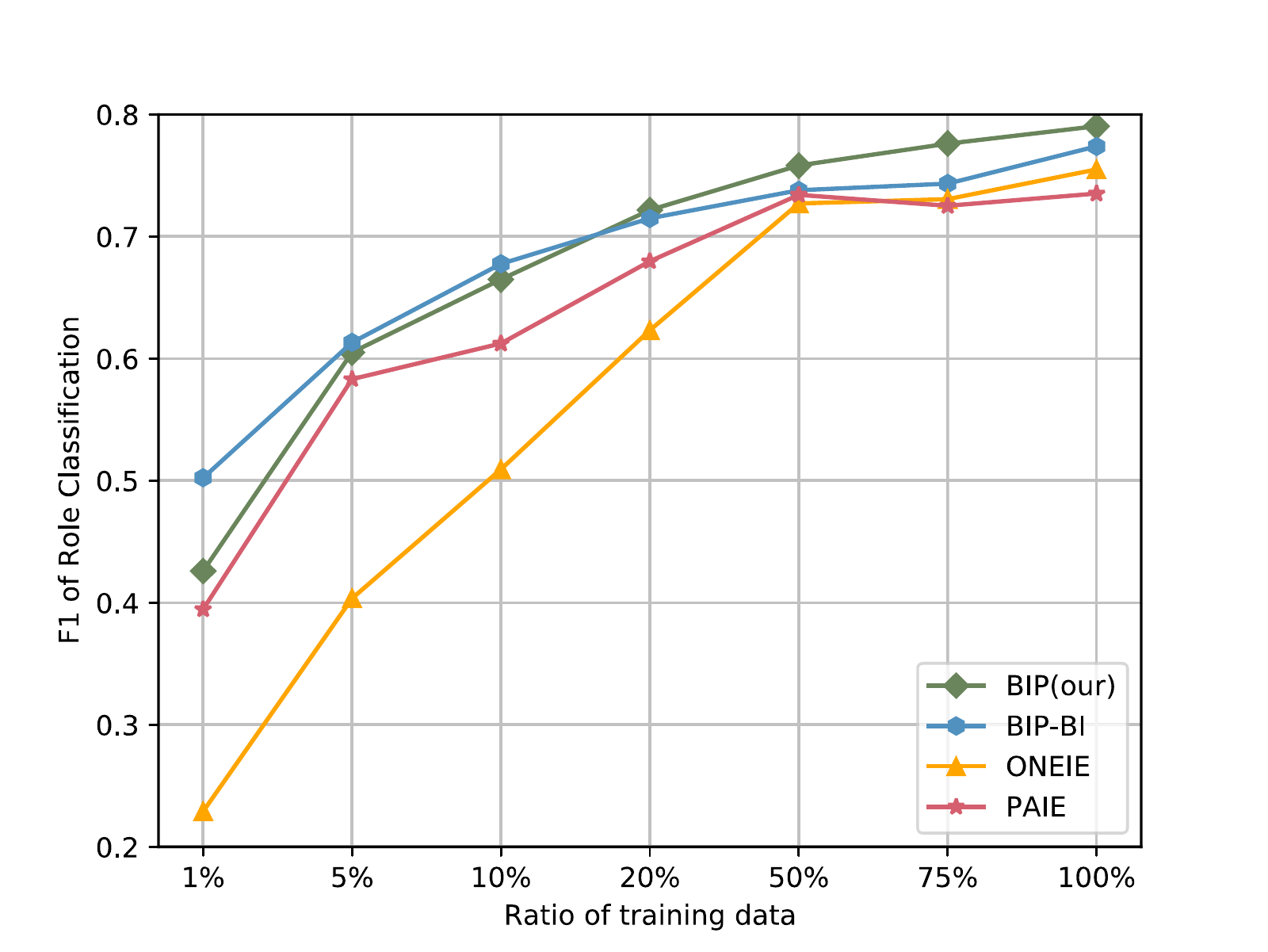}
    \caption{Performances against the ratio of training data, while ONEIE and PAIE are evaluated by flexible match.}\label{Fig:Data}
\end{figure}

Figure~\ref{Fig:Data} presents the performance of our BIP, BIP-BI and two state-of-the-art methods in both low-resource and high-resource data scenarios. We can observe that the variation of F1-score has a trend of rising with the increase of the training data. Comparing the fine-tuning method ONEIE, prompt-tuning methods BIP, BIP-BI and PAIE obviously improve the performance of role classification in low-resource data scenarios. This result shows that prompt-tuning methods can more effectively utilize the rich knowledge in PLMs than fine-tuning methods.

\par
Even using flexible match to evaluate the prompt-tuning method PAIE, our method BIP and BIP-BI achieve better performance in both low-resource and high-resource data scenarios. The main reason is that our method can make use of the entity information and predicted argument roles when constructing the template. We notice that the performance of BIP is worse than that of BIP-BI, when the ratio of training data is less than $20\%$. This is because when the number of training data is too small, the probability of argument roles being correctly predicted is low. If the bi-directional iterative strategy is adopted, the wrongly predicted argument roles will be used for template construction, which will further degrade the performance of EAE.

\subsection{Case Study}
\begin{table*}[t]
\centering
\begin{tabular}{|c|c|c|c|c|c|c|c|}
\hline
\multirow{2}{*}{Model} & \multirow{2}{*}{Template}  & \multicolumn{3}{c|}{Argument Identification} &\multicolumn{3}{c|}{Role Classification}\\ \cline{3-8}
                        & & P  & R & F1 & P  & R & F1 \\ \hline  \hline
\multirow{3}{*}{BIP(our)}    & Hard Template        & 78.17 & 86.40 & 82.08 & 75.26 & 83.19 & 79.03 \\
                        & Soft Template        & 80.63 & 82.75 & 81.67 & 77.49 & 79.53 & 78.50 \\
                        & Hard-Soft Template   & 77.15 & 82.46 & 79.72 & 74.15 & 79.24 & 76.61 \\  \hline  \hline
\multirow{3}{*}{BIP-BI}    & Hard Template        & 81.82 & 78.95 & 80.36 & 78.79 & 76.02 & 77.38 \\
                        & Soft Template        & 76.25 & 84.94 & 80.36 & 73.62 & 82.02 & 77.59 \\
                        & Hard-Soft Template   & 81.84 & 80.70 & 81.12 & 78.29 & 77.49 & 77.88 \\
\hline
\end{tabular}
\caption{Performance of different templates}\label{Tbl:Prompt}
\end{table*}

In order to showcase the effectiveness of our method BIP, we sample three sentences from the ACE 2005 English test dataset to compare the event argument extraction results by BIP, BIP(forward), BIP(backward), BIP-BI and BIP-SV methods.

\par
In Sentence 1 of Table~\ref{Tbl:Case2}, the method without the bi-directional iterative strategy BIP-BI can only identify the entity "\textbf{Bush}" as the "Entity" role. For the entity "\textbf{Putin}", the methods with the forward iterative prompt BIP, BIP(forward), BIP-SV can correctly classify it into the "Entity" role. This attributes to that the information that entity "\textbf{Bush}" is the "Entity" argument is introduced into the template construction of the entity "\textbf{Putin}". We also notice that "\textbf{Bush}" and "\textbf{Putin}" are both misclassified in the BIP(backward) method, where the error role information of "\textbf{Putin}" is passed to the classification of "\textbf{Bush}". In addition, for the entity "\textbf{[he, Erdogan]}" in Sentence 2, the method only with the forward iterative prompt BIP(forward) misclassifies the entity "\textbf{[Tomahawk, missiles]}" into the "Instrument" role. These results show that the argument roles of contextual entities can provide useful information for the role identification of the current entity. However, only considering argument interactions in one direction may degrade the performance of event argument extraction.

\par
In Sentence 3, the method BIP-SV misclassifies the entity "\textbf{SEC}" into the "Entity" role. For the human-written verbalizer of BIP-SV, the word "judge" is selected as the label word of role "Adjudicator". It is difficult to associate the entity "\textbf{SEC}" with the word "judge". In the semantical verbalizer, we use the text sequence "the entity doing the fining" to describe the semantic of "Adjudicator" role in the "Fine" event. Since the pre-trained language models can easily identify the entity "\textbf{SEC}" as "the entity doing the fining", the methods with semantical verbalizer can correctly identify the entity "\textbf{SEC}" as the "Adjudicator" role. The result verifies the effectiveness of our designed semantical verbalizer.

\subsection{Prompt Variants}\label{Subsec:Prompts}

In this section, we compare three different templates introduced in Section~\ref{Subsec:Templates} to investigate how different types of templates affect the performance of EAE. For the BIP-BI method, the performances of hard template, soft template and hard-soft template are comparable. Since the hard-soft template combines the manual knowledge and learnable virtual tokens, it achieves the best performance. However, the hard-soft template performs worst for the BIP method. Unlike the BIP-BI method which only considers event trigger and current entity information, BIP introduces the predicted argument role information into the template. Therefore, there are so many learnable pseudo tokens in the hard-soft template, resulting in poor performance.

\section{Conclusion and Future Work}\label{Sec:Conclusion}
In this paper, we regard event argument extraction as a cloze-style task and propose a bi-directional iterative prompt-tuning method to address this task. The bi-directional iterative prompt-tuning method contains a forward iterative prompt and a backward iterative prompt, which predict the argument role of each entity in a left-to-right and right-to-left manner respectively. For the template construction in each prompt, the predicted argument role information is introduced to capture argument interactions. In addition, a novel semantical verbalizer is designed based on the semantic of the argument role. And three kinds of templates are designed and discussed. Experiment results have shown the effectiveness of our method in both high-resource and low-resource data scenarios. In the future work, we are interested in the joint prompt-tuning method of event detection and event argument extraction.

\section*{Limitations}
\begin{itemize}
  \item As the entity information is necessary to model event argument extraction as a cloze-style task, our method is not suitable for the situation that entities are not provided.
  \item Comparing with the methods that predict argument roles simultaneously, the speed of our method is slower due to that it predicts the argument role of each entity one by one.
\end{itemize}

\bibliography{anthology,custom}
\bibliographystyle{acl_natbib}

\appendix
\section{Verbalizer} \label{sec:verbalizer}
\subsection{Semantical Verbalizer}
For our designed semantical verbalizer, an argument role that participates in multiple types of events is divided into multiple argument roles that are specific to event types. For each new argument role, we use a virtual word to represent the role and initialize the representation of the virtual word with the semantic of the argument role. Table~\ref{Tbl:Verbalizer} shows the redefined argument role types, and the semantic description and virtual label word of each argument role type.

\begin{table*}[t]
\scriptsize
\centering
\begin{tabular}{|c|c|c|}
\hline
Redefined Argument Role Label & Semantic Description & Virtual Label Word\\ \hline
Event:None   & the entity that is irrelevant to the event & Event:None \\ \hline
Event:Place   & the place where the event takes place &  Event:Place\\ \hline
Be-Born:Person   & the person who is born &  Be-Born:Person\\ \hline
Marry:Person   & the person who are married &  Marry:Person\\ \hline
Divorce:Person   & the person who are divorced &  Divorce:Person\\ \hline
Injure:Agent   & the one that enacts the harm &  Injure:Agent\\ \hline
Injure:Victim   & the harmed person &  Injure:Victim\\ \hline
Injure:Instrument   & the device used to inflict the harm &  Injure:Instrument\\ \hline
Die:Agent   & the killer &  Die:Agent\\ \hline
Die:Victim   & the person who died &  Die:Victim\\ \hline
Die:Instrument   & the device used to kill &  Die:Instrument\\ \hline
Transport:Agent   & the agent responsible for the transport event & Transport:Agent \\ \hline
Transport:Artifact   & the person doing the traveling or the artifact being transported &  Transport:Artifact\\ \hline
Transport:Vehicle   & the vehicle used to transport the person or artifact &  Transport:Vehicle\\ \hline
Transport:Origin   & the place where the transporting originated &  Transport:Origin\\ \hline
Transport:Destination   & the place where the transporting is directed &  Transport:Destination\\ \hline
Transfer-Ownership:Buyer   & the buying agent &  Transfer-Ownership:Buyer\\ \hline
Transfer-Ownership:Seller   & the selling agent &  Transfer-Ownership:Seller\\ \hline
Transfer-Ownership:Beneficiary   & the agent that benefits from the transaction &  Transfer-Ownership:Beneficiary\\ \hline
Transfer-Ownership:Artifact   & the item or organization that was bought or sold &  Transfer-Ownership:Artifact\\ \hline
Transfer-Money:Giver   & the donating agent &  Transfer-Money:Giver\\ \hline
Transfer-Money:Recipient   & the recipient agent &  Transfer-Money:Recipient\\ \hline
Transfer-Money:Beneficiary   & the agent that benefits from the transfer & Transfer-Money:Beneficiary \\ \hline
Start-Org:Agent   & the founder &  Start-Org:Agent\\ \hline
Start-Org:Org   & the organization that is started &  Start-Org:Org\\ \hline
Merge-Org:Org   & the organizations that are merged &  Merge-Org:Org\\ \hline
Declare-Bankruptcy:Org   & the organization declaring bankruptcy &  Declare-Bankruptcy:Org\\ \hline
End-Org:Org   & the organization that is ended &  End-Org:Org\\ \hline
Attack:Attacker   & the attacking agent &  Attack:Attacker\\ \hline
Attack:Target   & the target of the attack &  Attack:Target\\ \hline
Attack:Instrument   & the instrument used in the attack &  Attack:Instrument\\ \hline
Demonstrate:Entity   & the demonstrating agent &  Demonstrate:Entity\\ \hline
Meet:Entity   & the agents who are meeting &  Meet:Entity\\ \hline
Phone-Write:Entity   & the communicating agent & Phone-Write:Entity \\ \hline
Start-Position:Person   & the employee &  Start-Position:Person\\ \hline
Start-Position:Entity   & the employer &  Start-Position:Entity\\ \hline
End-Position:Person   & the employee &  End-Position:Person\\ \hline
End-Position:Entity   & the employer &  End-Position:Entity\\ \hline
Elect:Person   & the person elected &  Elect:Person\\ \hline
Elect:Entity   & the voting agent &  Elect:Entity\\ \hline
Nominate:Person   & the person nominated &  Nominate:Person\\ \hline
Nominate:Agent   & the nominating agent &  Nominate:Agent\\ \hline
Arrest-Jail:Person   & the person who is jailed or arrested &  Arrest-Jail:Person\\ \hline
Arrest-Jail:Agent   & the jailer or the arresting agent &  Arrest-Jail:Agent\\ \hline
Release-Parole:Person   & the person who is released &  Release-Parole:Person\\ \hline
Release-Parole:Entity   & the former captor agent &  Release-Parole:Entity\\ \hline
Trial-Hearing:Defendant   & the agent on trial & Trial-Hearing:Defendant \\ \hline
Trial-Hearing:Prosecutor   & the prosecuting agent &  Trial-Hearing:Prosecutor\\ \hline
Trial-Hearing:Adjudicator   & the judge or court &  Trial-Hearing:Adjudicator\\ \hline
Charge-Indict:Defendant   & the agent that is indicted &  Charge-Indict:Defendant\\ \hline
Charge-Indict:Prosecutor   & the agent bringing charges or executing the indictment &  Charge-Indict:Prosecutor\\ \hline
Sue:Plaintiff   & the suing agent &  Sue:Plaintiff\\ \hline
Sue:Defendant   & the agent being sued &  Sue:Defendant\\ \hline
Sue:Adjudicator   & the judge or court &  Sue:Adjudicator\\ \hline
Convict:Defendant   & the convicted agent &  Convict:Defendant\\ \hline
Convict:Adjudicator   & the judge or court &  Convict:Adjudicator\\ \hline
Sentence:Defendant   & the agent who is sentenced &  Sentence:Defendant\\ \hline
Sentence:Adjudicator   & the judge or court & Sentence:Adjudicator \\ \hline
Fine:Entity   & the entity that was fined &  Fine:Entity\\ \hline
Fine:Adjudicator   & the entity doing the fining &  Fine:Adjudicator\\ \hline
Execute:Person   & the person executed &  Execute:Person\\ \hline
Execute:Agent   & the agent responsible for carrying out the execution &  Execute:Agent\\ \hline
Extradite:Person   & the person being extradited &  Extradite:Person\\ \hline
Extradite:Agent   & the extraditing agent &  Extradite:Agent\\ \hline
Extradite:Origin   & the original location of the person being extradited &  Extradite:Origin\\ \hline
Extradite:Destination   & the place where the person is extradited to &  Extradite:Destination\\ \hline
Acquit:Defendant   & the agent being acquitted &  Acquit:Defendant\\ \hline
Acquit:Adjudicator   & the judge or court &  Acquit:Adjudicator\\ \hline
Pardon:Defendant   & the agent being pardoned &  Pardon:Defendant\\ \hline
Pardon:Adjudicator   & the state official who does the pardoning &  Pardon:Adjudicator\\ \hline
Appeal:Defendant   & the defendant &  Appeal:Defendant\\ \hline
Appeal:Adjudicator   & the judge or court &  Appeal:Adjudicator\\ \hline
Appeal:Plaintiff   & the appealing agent &  Appeal:Plaintiff\\ \hline
\end{tabular}
\caption{Label words of the human-written verbalizer.}\label{Tbl:SemanVer}
\end{table*}

\subsection{Human-written Verbalizer}
For the human-written verbalizer, we assign a label word to each argument role. Table~\ref{Tbl:Verbalizer} lists the label word of each argument role.

\begin{table}[t]
\centering
\begin{tabular}{|c|c|}
\hline
Argument Role Label & Label Word \\ \hline
None   & none \\ \hline
Person   & person \\ \hline
Place   & place \\ \hline
Buyer   & buyer \\ \hline
Seller   & seller \\ \hline
Beneficiary   & beneficiary \\ \hline
Artifact   & artifact \\ \hline
Origin   & origin \\ \hline
Destination   & destination \\ \hline
Giver   & donor \\ \hline
Recipient   & recipient \\ \hline
Org   & organization \\ \hline
Agent   & agent \\ \hline
Victim   & victim \\ \hline
Instrument   & instrument \\ \hline
Entity   & entity \\ \hline
Attacker   & attacker \\ \hline
Target   & target \\ \hline
Defendant   & defendant \\ \hline
Adjudicator   & judge \\ \hline
Prosecutor   & prosecutor \\ \hline
Plaintiff   & plaintiff \\ \hline
Vehicle   & vehicle \\ \hline
\hline
\end{tabular}
\caption{Label words of the human-written verbalizer.}\label{Tbl:Verbalizer}
\end{table}

\section{Templates}
\label{sec:templates}
For our designed templates, each entity (event) type is converted into a human-understandable
text span, so as to take full advantage of event type label and entity type label information. Table~\ref{Tbl:Enttype} and ~\ref{Tbl:Eventtype} list all text spans of entity types and event types.

\begin{table}[t]
\centering
\begin{tabular}{|c|c|}
\hline
Entity Type & Text Span \\ \hline
FAC   & facility \\ \hline
ORG   & organization \\ \hline
GPE   & geographical or political entity \\ \hline
PER   & person \\ \hline
VEH   & vehicle \\ \hline
WEA   & weapon \\ \hline
LOC   & location \\ \hline
\hline
\end{tabular}
\caption{Text spans of entity types.}\label{Tbl:Enttype}
\end{table}

\begin{table}[t]
\centering
\small
\begin{tabular}{|c|c|}
\hline
Event Type & Text Span \\ \hline
Transport   & transport \\ \hline
Elect   & election \\ \hline
Start-Position   & employment \\ \hline
End-Position   & dimission \\ \hline
Attack   & attack \\ \hline
Meet   & meeting \\ \hline
Marry   & marriage \\ \hline
Transfer-Money   & money transfer \\ \hline
Demonstrate   & demonstration \\ \hline
End-Org   & collapse \\ \hline
Sue   & prosecution \\ \hline
Injure   & injury \\ \hline
Die   & death \\ \hline
Arrest-Jail   & arrest or jail \\ \hline
Phone-Write   & written or telephone communication \\ \hline
Transfer-Ownership   & ownership transfer \\ \hline
Start-Org   & organization founding \\ \hline
Execute   & execution \\ \hline
Trial-Hearing   & trial or hearing \\ \hline
Be-Born   & birth \\ \hline
Charge-Indict   & charge or indict \\ \hline
Sentence   & sentence \\ \hline
Declare-Bankruptcy   & bankruptcy \\ \hline
Release-Parole   & release or parole \\ \hline
Fine   & fine \\ \hline
Pardon   & pardon \\ \hline
Appeal   & appeal \\ \hline
Extradite   & extradition \\ \hline
Divorce   & divorce \\ \hline
Merge-Org   & organization merger \\ \hline
Acquit   & acquittal \\ \hline
Nominate   & nomination \\ \hline
Convict  & conviction \\ \hline
\end{tabular}
\caption{Text spans of event types.}\label{Tbl:Eventtype}
\end{table}

\end{document}